\documentclass[10pt,twocolumn,letterpaper]{article}

\usepackage[pagenumbers]{cvpr} 

\usepackage{graphicx}
\usepackage{amsmath}
\usepackage{amssymb}
\usepackage{booktabs}

\usepackage{enumitem}
\usepackage{multirow}
\usepackage{bbding}
\usepackage{indentfirst}
\usepackage[normalem]{ulem}
\usepackage{changepage}
\usepackage{multicol}

\usepackage[pagebackref,breaklinks,colorlinks]{hyperref}

\usepackage[capitalize]{cleveref}
\crefname{section}{Sec.}{Secs.}
\Crefname{section}{Section}{Sections}
\Crefname{table}{Table}{Tables}
\crefname{table}{Tab.}{Tabs.}

\def\cvprPaperID{9568} 
\def\confName{CVPR}
\def\confYear{2023}

\begin{document}

\title{Diversity-Measurable Anomaly Detection}

\author{Wenrui Liu$^{1,2}$, Hong Chang$^{1,2}$, Bingpeng Ma$^{2}$, Shiguang Shan$^{1,2}$, Xilin Chen$^{1,2}$\\
$^1$Institute of Computing Technology, Chinese Academy of Sciences, China\\
$^2$University of Chinese Academy of Sciences, China\\
{\tt\small wenrui.liu@vipl.ict.ac.cn, $\left\{\right.$changhong, sgshan, xlchen$\left.\right\}$@ict.ac.cn, bpma@ucas.ac.cn}
}
\maketitle

\begin{abstract}

    Reconstruction-based anomaly detection models achieve their purpose by suppressing the generalization ability for anomaly. However, diverse normal patterns are consequently not well reconstructed as well. Although some efforts have been made to alleviate this problem by modeling sample diversity, they suffer from shortcut learning due to undesired transmission of abnormal information. In this paper, to better handle the tradeoff problem, we propose \textbf{Diversity-Measurable Anomaly Detection (DMAD)} framework to enhance reconstruction diversity while avoid the undesired generalization on anomalies. To this end, we design \textbf{Pyramid Deformation Module (PDM)}, which models diverse normals and measures the severity of anomaly by estimating multi-scale deformation fields from reconstructed reference to original input. Integrated with an information compression module, PDM essentially decouples deformation from prototypical embedding and makes the final anomaly score more reliable. Experimental results on both surveillance videos and industrial images demonstrate the effectiveness of our method. In addition, DMAD works equally well in front of contaminated data and anomaly-like normal samples.
\end{abstract}

\section{Introduction}
\label{sec:1}
Visual anomaly detection is a fundamental and important problem in computer vision community, with wide applications in video surveillance and industrial inspection. It aims to detect outliers from seen classes and novel patterns from unseen classes. This task is very challenging because abnormal data is diversely distributed and expensive to collect. So we have to construct models based on only normal samples under unsupervised setting, targeting at high discrimination between normal and abnormal samples.

During the past decade, reconstruction-based methods have achieved great progress in anomaly detection. These methods use Autoencoders (AEs) \cite{convae2d,convae3d,convaelstm,memae,mnad,mpn,hf2vad} or Generative Adversarial Networks (GANs) \cite{framepred,gn,anogan} to reconstruct the normal counterparts from any input images or video frames. AE-based methods firstly compress the inputs to discard the information beyond normal prototypes, and then decode the embedding to reconstruct the inputs. According to the estimated reconstruction error, the anomalies can be detected.
\begin{figure}[t]
  \centering
  \includegraphics[width=0.65\linewidth]{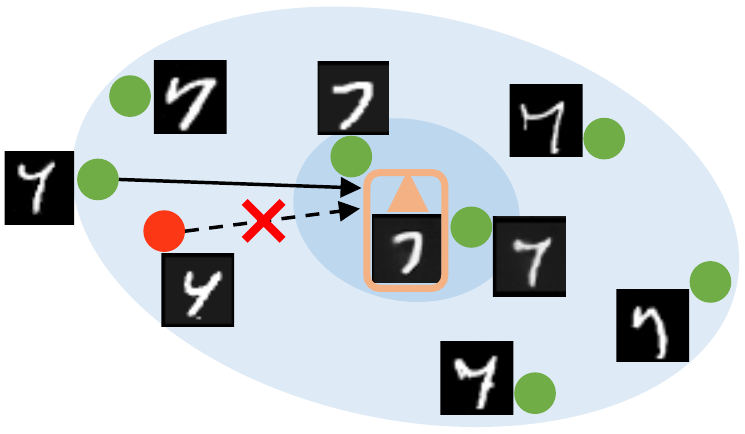}
  \caption{Illustration of difficulty in anomaly detection in MNIST dataset. The prototype is indicated by orange triangle and the anomaly by red point. In this case, the anomaly can hardly be detected based on reconstruction error or distance in high-dimensional feature space. Our solution is illustrated in Fig. 2.}
\vspace{-0.1em}
  \label{fig:1}
\end{figure}

\begin{figure}
  \centering
  \includegraphics[width=1.0\linewidth]{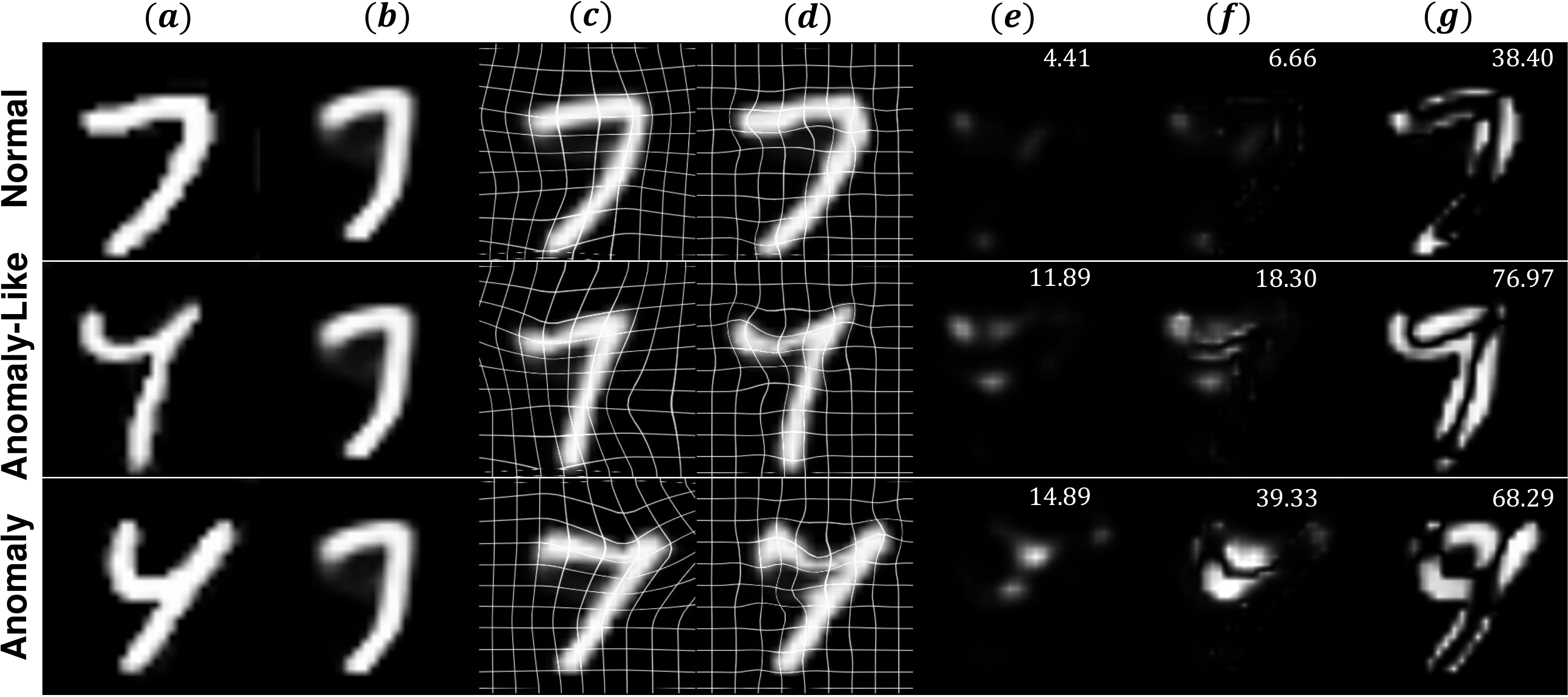}
  \caption{Illustration of our diversity-measurable method in addressing the detection difficulty. Numbers in white are anomaly scores. \textbf{a)} Original input; \textbf{b)} Reconstructed reference; \textbf{c)} Coarse deformation; \textbf{d)} Fine deformation; \textbf{e)} Measurement of diversity\protect\footnotemark[1]; \textbf{f)} Deformation-augmented error map assigns lower anomaly score to the anomaly-like sample than the true anomaly; \textbf{g)} Pixel-wise reconstruction error yields incorrect anomaly scores.}
  \vspace{-0.1em}
  \label{fig:2}
\end{figure}

However, the performance of reconstruction-based methods for anomaly detection has long been limited by a tough problem, \ie \textit{the tradeoff between reconstructing diverse normals and detecting unknown anomalies}.
In order to discriminate anomalies more easily, previous works \cite{memae} imposes more constraints to suppress abnormal information during autoencoding, which leads to high reconstruction error for diverse normal instances. For example, in \cref{fig:1,fig:2}\textbf{g}, the severely deformed normal (a.k.a. anomaly-like) sample ``7'' has even higher error than the abnormal sample ``4''. To better reconstruct diverse normals, each query vector correspond to multiple prototypes in the memory, which may be combined into abnormal embedding even if abnormal projection is far away from the prototype. As a consequence, anomalies that distribute in low likelihood area between prototypical embedding are difficult to identify from diverse normals. MNAD\cite{mnad} introduces skip-connection for diverse reconstruction and additional constraints to get round the incorrect combination problem. But the latter forces model transmit more unrestrained information with abnormal part by skip-connection, resulting in shortcut learning and undesired reconstruction of anomalies.

A key to address the above tradeoff problem is to find a proper measurement of diversity that normal and abnormal samples have, which is positively correlated with the severity of anomaly. With such a measure, we do not need to fight against imperfect reconstruction of normals or undesired reconstruction of anomalies, because anomalies can be detected more accurately by the diversity measure together with the reconstruction error. Note that pixel-wise reconstruction error is not an ideal measurement of diversity, because the high-error region often confuses anomalies with diverse normals, \eg normals with structural deformation and anomalies with colors close to the background may yield unreliable reconstruction error.

In this paper, we propose a \textit{\textbf{Diversity-Measurable Anomaly Detection (DMAD)}} framework to enhance the measurability of reconstruction diversity so as to measure abnormality more accurately. Our basic idea is to decouple the reconstruction into compact representation of prototypical normals and measurable deformations of more diverse normals and anomalies. The under-estimated reconstruction error can be compensated by the diversity, which can be properly measured. To this end, the DMAD framework includes a \textit{\textbf{Pyramid Deformation Module (PDM)}} to model and measure the diversity and an Information Compression Module (ICM) to learn the prototypical normal patterns.

\footnotetext[1]{In this case, we only count fine deformation because deformations in position and angle are considered as normal. In real-world experiments, we consider both coarse and fine deformations.}

Inspired by \cite{stn,dcn}, we assume anomalies (\eg in video surveillance) can be represented as significant deformation of appearances, including positional changes and fine motions. In contrast, diverse normal samples can be represented as weaker deformations thus easily distinguished from the abnormal ones. Therefore, we design PDM to model the diversity of normals as well as the severity of anomalies. More specifically, PDM learns hierarchical two-dimensional deformation fields (\cref{fig:2}\textbf{c},\textbf{d}) that describe the pixel-level transformation direction and distance from reference (\cref{fig:2}\textbf{b}, which is reconstructed from prototypes in memory) to original input. In ICM, we learns compressed representation as sparse prototypes. As a result, a single memory item is enough to represent each normal cluster. This is more compact than other memory-based works which require multiple memory items. Integrating PDM with ICM, DMAD essentially decouples the deformation information (\cref{fig:2}\textbf{e}) from class prototypes and makes the final anomaly score more discriminative (\cref{fig:2}\textbf{f}). 

We evaluate our anomaly detection framework in scenarios of video surveillance and industrial defect detection. To apply DMAD in the latter scenario, we propose a variant of PDM, PPDM, to deal with the false positive issue in texture reconstruction. Extensive experimental results verify the efficacy of our approach. Moreover, our method works well even in front of contaminated data and anomaly-like normals. The main contributions of our work are as follows:
\begin{itemize}[leftmargin= 1 em, topsep= 2.pt, parsep= 0pt, itemsep=2.pt, partopsep=0pt]
    \item
    We introduce diversity-measurable anomaly detection framework which allows reconstruction-based models to achieve better tradeoff between reconstructing diverse normals and detecting unknown anomalies.
    \item
    We propose pyramid deformation module to implement diversity measurement, in which the deformation information is explicitly separated from compact class prototypes and the resulting diversity measure is positively correlated to abnormality.
    \item
    Our approach outperforms previous works on video anomaly detection and industrial defect detection, and works well in front of contaminated data and anomaly-like normals, demonstrating its broad suitability and robustness.
\end{itemize}

\section{Related Work}
    \textbf{Anomaly detection.}
    Reconstruction-based methods model the distribution of normal data and assign anomalies with high reconstruction error, because models trained with only normal data cannot reconstruct anomalies. Some works use autoencoder to detect anomalies, such as convolutional autoencoder \cite{convae2d} and the variants \cite{convae3d,convaelstm}. Other approaches introduce additional constraints or memory to make the model more discriminative. For example, sparse coding \cite{onlinesparsecode,timesparse} reduces representation redundancy with regularization; memory-augmented autoencoder (MemAE) \cite{memae} memorizes the normal patterns appearing in training dataset with external memory bank; variational autoencoder \cite{betavae,vqvae} assumes a prior distribution of normal data to constrain the nonlinear representation capacity; HF$^2$VAD \cite{hf2vad} uses CVAE \cite{cvae} to capture the correlation among motions. Frame prediction \cite{framepred} assumes that abnormal samples in videos cannot be represented by past frames that do not contain unseen information and forces the model to encode changes among different frames. In addition, autoencoders can also be combined with external object detectors \cite{hf2vad,ssml,ocae} to capture background-invariant appearance.

    Although these methods generally work well, they often have difficulty in discriminating abnormal samples from anomaly-like normals due to the tradeoff between reconstruction and discrimination. In our framework, pyramid deformation module and information compression module are leveraged to address the trade-off problem, significantly improving the performance in anomaly detection.

    \textbf{Memory network.}
    Generative models map continuously in feature space and they may assign higher probabilities to anomalies than normal ones \cite{dontknow}. Recent research \cite{memae,mnad,hf2vad,daad} has explored the application of discrete external memory to generate seen normal patterns even if the input is abnormal. MemAE \cite{memae} proposes a memory-augmented autoencoder, which uses information from encoder as a query and obtains a normal pattern retrieved from memory module. MNAD \cite{mnad} introduces skip-connection to alleviate the problem that diverse normal patterns may yield high reconstruction errors. HF$^2$VAD \cite{hf2vad} extends the memory module to multi-level memory, and uses additional estimators \cite{cascadercnn, flownet} to explicitly model the motion information and filter the noise from background. DAAD \cite{daad} uses block-wise memory to increase the specificity of memory.

    In most existing works, the memory module outputs a linear combination of memory items which may lead to undesired reconstruction of normal-like anomalies. And they do not explicitly consider that instances corresponding to the same memory grid may be located at different positions of the receptive field. In our approach, we alleviate these problems via compressing embedding into a single memory item to ensure that the output is absolutely normal.

    \textbf{Transformation modeling.}
    The transformation between video frames is a key clue for anomaly detection. Some methods \cite{hf2vad,framepred,ssml,ocae} use external object detectors or optical-flow estimators to model the motion information implicitly or explicitly. Among the transformation estimators, optical-flow estimation \cite{flownet} is widely studied, which aims to estimate the pixel-wise motions between consecutive frames. In addition, correspondence modeling between pair-wise images is also applied to 2D geometric matching \cite{match2d} and 3D image registration \cite{reg3d} to obtain image deformation fields. STN \cite{stn} learns to transform the original images themselves to benefit identification algorithm. RegAD \cite{regad} uses feature-level affine registration to relocate features without changing the embeddings.

    However, pairwise transformation modeling only focus on pattern changes thus cannot detect static anomalies. And pretrained transformation estimators introduce generalization error in different scenarios. In contrast, we address these problems by separating and measuring the transformation from prototypical memory in end-to-end training.
    
    \begin{figure}
  \centering
  \includegraphics[width=1\linewidth]{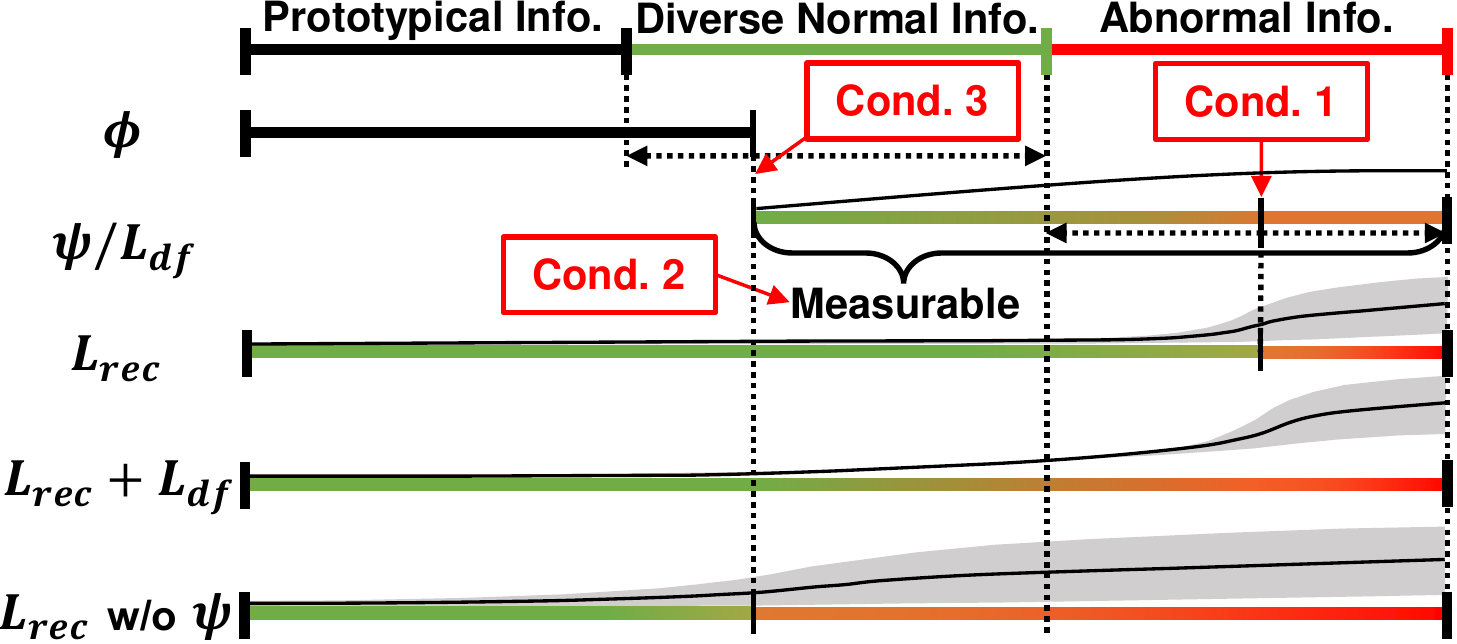}
  \vspace{-1.em}
  \caption{Three conditions for DMAD framework. The colors of lines represent the measurements of diversity (severity of anomaly); The gray area represents uncertainty in the measure; Double dashed arrows indicate possible ranges of the boundary. The last two lines denote  DMAD and traditional reconstruction-based method respectively.}
  \vspace{-0.1em}
  \label{fig:cond}
\end{figure}

\begin{figure*}
  \centering
  \includegraphics[width=0.99\linewidth]{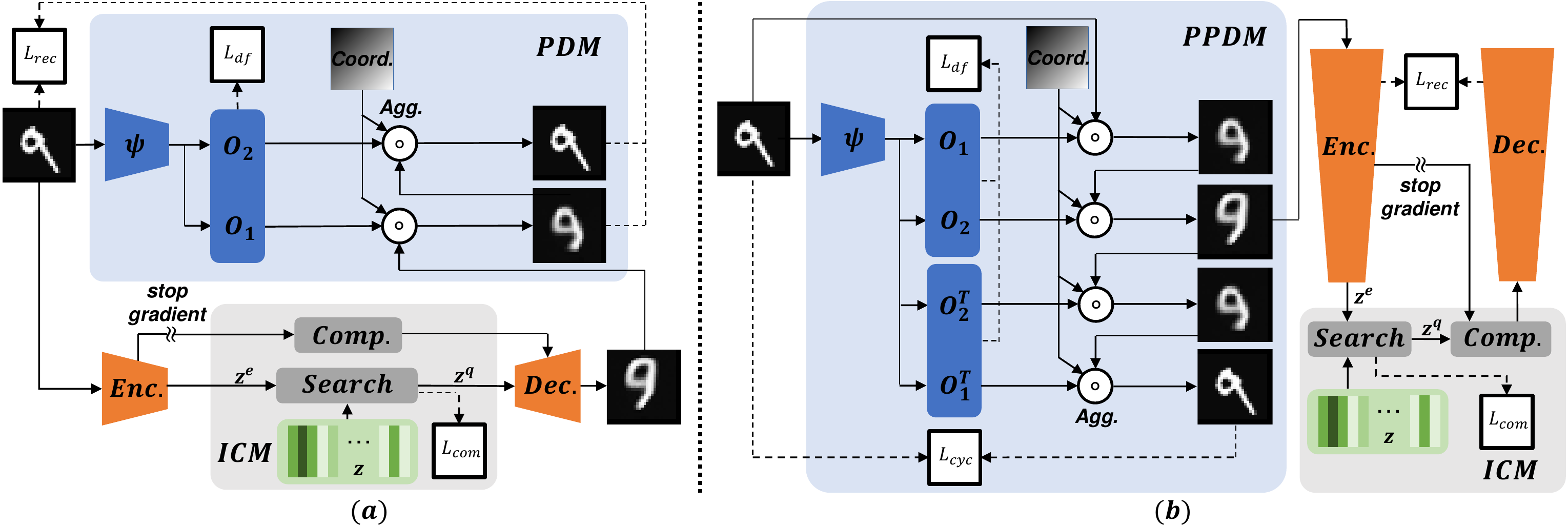}
  \vspace{-0.3em}
  \caption{Two versions of diversity-measurable anomaly detection framework. Multi-scale pyramid deformation fields are estimated as $\mathbf{O}$ and the reverse processes are $\mathbf{O}^T$. \textbf{a)} PDM version computes forward deformation $\mathbf{O}$ after reconstruction. \textbf{b)} PPDM version employs cycle-consistent forward-backward deformations and the forward deformation is applied on the input.
 }
  \vspace{-0.1em}
  \label{fig:3}
\end{figure*}

\section{Diversity-Measurable Anomaly Detection}
    In this section, we first analyze the objective of reconstruction-based anomaly detection and propose a diversity-measurable framework to address the tradeoff problem in existing works. Then, we introduce information compression module and pyramid deformation module (PDM) as an implementation of the framework. Finally, we explain the training and inference process and how to apply the framework to defect detection with a variant of PDM.
   
    \subsection{The framework}
        \label{framework}
        Given input $x$, autoencoder-based methods aim to model normal data distribution by minimizing the following reconstruction loss ($\Vert \cdot \Vert_2$ is just one type of reconstruction loss):
        \begin{equation}
        L = {\Vert x - g(\phi(f(x), z))\Vert} _2 + \gamma _1 R_1(\phi),
        \end{equation}
        \noindent
        with respect to encoder $f(\cdot)$, decoder $g(\cdot)$, latent variable $z$ (referring to discrete memory items in these cases) and a constrained feature mapping function $\phi(\cdot)$ corresponding to its constraint $R_1(\phi)$. Skip-connection and concatenation $[\cdot,\cdot]$ are introduced to generate diverse normal patterns\cite{mnad}:
        \begin{equation}
        L = {\Vert x - g([\phi(f(x), z), f(x)])\Vert} _2 + \gamma _1 R_1(\phi).
        \end{equation}
        However, due to diverse data distribution, previous methods have to face the conflict between representing diverse normals and detecting anomalies. The intrinsic reason lies in that the encoding of diversity $[\cdot, f(x)]$ contains redundant information that cannot be measured accurately.
         
    In this work, we propose a principled framework, \textit{\textbf{D}iversity-\textbf{M}easurable \textbf{A}nomaly \textbf{D}etection (\textbf{DMAD})}, to alleviate the conflict. 
    The basic idea is to restrict the anomaly information transmitted to $g(\cdot)$ while measuring and modeling the diversity of the remaining part. To this end, we design information compression module $\phi(\cdot)$ and diversity-aware module $\psi(\cdot)$ under DMAD framework:
        \begin{equation}
        L = {\Vert x - g(\phi(f(x), z)) \circ \psi(x)\Vert} _2 + \gamma _1 R_1(\phi) + \gamma _2 R_2(\psi),
        \label{eq:dmad}
        \end{equation}
        \noindent
        where $\circ$ refers to aggregation operator. With appropriate design of $\phi(\cdot)$, $\psi(\cdot)$ and the constraints, optimization of the reconstruction loss can improve the compactness of feature embedding. So that diverse representations are mapped via $\phi(\cdot)$ to compact prototypes in the memory. The diversity of input $x$ relative to its reconstruction is represented by $\psi(\cdot)$. The under-estimated reconstruction error can be compensated by the diversity measured in $\psi(\cdot)$, a key factor contributing to accurate anomaly scores.
        
        This framework can achieve our target on the premise that the following conditions (\cref{fig:cond}) are met: 1. $\psi(\cdot)$ can learn \textit{all} diverse information from prototypical patterns to any normal inputs to ensure normal samples do not yield high anomaly score; 2. Deformation generated by $\psi(\cdot)$ is positively correlated to diversity measure; 3. Prototypical information about $x$ represented by $\psi(\cdot)$ needs to be minimized. In the following subsections, we explain how to design modules to fulfill these conditions.

 \subsection{Information compression module}
            According to \cite{vqvae}, we adapt VQ-Layer as an information compression module to learn $\phi(\cdot)$ given embedding $f(x)\in R^{D \times H^\prime \times W^\prime}$ as a query $z^e = f(x)$ and memory $z \in R^{D \times N}$. Then, we quantize $z^e$ into a single-memory feature cube $z^q \in R^{D \times H^\prime \times W^\prime}$ by seeking for memory item with minimum $L2$ distance (``\textit{Search}'' in \cref{fig:3}):
            \begin{equation}
            z^q_{h, w} = {argmin}_{z_n} {\Vert z^e_{h, w} - z_n \Vert} _2,
            \end{equation}
            where $z_n$ is the $n^{th}$ memory item, $h \in \left\{1, \cdots, H^\prime \right\}, w \in \left\{1, \cdots, W^\prime \right\}$ indicates the same location in both $z^q$ and $z^e$. 
            The compression loss $L_{com}$ with stop-gradient operator $SG(\cdot)$ that updates its parameters separately\cite{vqvae} is combined by a hyperparameter $\beta$:
            \begin{equation}
            L_{com} = {\Vert SG(z^e) - z^q\Vert} _2 + \beta {\Vert z^e - SG(z^q)\Vert} _2.
            \end{equation}

            Skip-connection with low information capacity (\cref{fig:3} ``\textit{Comp.}'') is also allowed to further improve the reconstruction quality without bringing excessive generalization (\ie Conv-Layer with stop-gradient operator for intermediate features whose reduction factor is 16 or larger).
            
    \subsection{Pyramid deformation module}
        We categorize the unknown anomalies into the following three types: unseen class (\eg novel objects), global anomaly (\eg unexpected movement) and local anomaly (\eg strange behavior and workpiece damage) of seen class. The unseen class is easy to be detected based on reconstruction result, but the latter two types are usually confused with diverse normals. To discriminate these anomalies from normal ones, we represent the diversity using measurable deformation between reconstructed reference and original input, so that slight deformations occur in normals while drastic deformations occur in anomalies. 
        
        Inspired by STN and DCN \cite{stn,dcn}, we introduce \textit{\textbf{P}yramid \textbf{D}eformation \textbf{M}odule (\textbf{PDM})} which explicitly learn deformation fields with hierarchical scales to model the motion, behavior and defect of different anomaly types, as shown in \cref{fig:3}a. Specifically, after feature extraction, $\psi(\cdot)$ uses $K$-heads to compute offsets $\mathbf{O}=\left\{O_1,\cdots, O_K\right\}$, corresponding to $K$ coarse-to-fine deformations: 
        \begin{equation}
         \psi(x) = Up(h(PE(x)))=\mathbf{O},
        \end{equation}
        where $PE(\cdot)$ is positional embedding operator \cite{coordconv}, $h:R^{ C \times H \times W}\rightarrow R^{ 2 \times \left\{H_1 \times W_1,\cdots, H_K \times W_K\right\}}$ is the deformation estimator that generates offset vectors, $Up(\cdot)$ is upsampling function that resizes the outputs of $K$-heads to the same size with the original image. In our experiments, we set $K = 2$, with $O_1$ used to estimate coarse deformation (\eg corresponding to the position of pedestrians or the placement of workpieces) and $O_2$ used to estimate fine deformation (\eg corresponding to pedestrian behaviors or workpiece details).
            
        Considering that quantized embedding without positional information may lead to inaccurate reconstruction, we also introduce position embedding operator for the decoder $g(\cdot)$. We then aggregate $\mathbf{O}$ onto the reconstructed reference $g(PE(z^q))$, and obtain $\tilde{x}_k (k=1,\ldots,K)$ which is calibrated by the $k^{th}$ layer of deformation fields:
        \begin{equation}
          \tilde{x}_k = g(PE(z^q)) \circ O_1 \cdots \circ O_k,
        \end{equation}
        where $ \circ $ is grid-sampling function with a reference coordinate in this implementation (``\textit{Agg.}'' and ``\textit{Coord.}'' in \cref{fig:3}). However, minimizing the unconstrained reconstruction loss with respect to $\psi(x)$ may cause degenerate solution of the encoder $f(\cdot)$. To address this problem, we add constraint using smoothness loss via gradient operation and strength loss as:
        \begin{equation}
        L_{df} = \sum_k {\Vert \nabla O_k\Vert} _1 + {\Vert O_k\Vert} _2.
        \end{equation}
        
    \subsection{Foreground-background selection}
        Storing the background information in memory will break the compactness of embedding and require plenty of memory items. Besides, the deformation estimation should not be applied to the background. Some approaches use external estimators to remove interference from background, but the generalization in different scenarios cannot be guaranteed and extra noise will inevitably be introduced. Benefiting from the strong prior of fixed-view videos, we model the background with a learnable template $x_{bg}$ and generate a binary mask to indicate whether a pixel belongs to foreground or background with $f_m(\cdot)$. The final reconstruction $\hat{x}_k$ of $k^{th}$ head is:
        \begin{equation}
          \hat{x}_k = f_m(x)  \tilde{x}_k + (1 - f_m(x))  x_{bg}.
          \label{eq:mask}
        \end{equation}

    \subsection{Training and inference}
        \textbf{Training phase.}
        Once we obtain the reconstruction $\hat{x}_k$, we can calculate reconstruction loss $L_{rec}$ as:
        \begin{equation}
            L_{rec} = \sum _k Dis(x, \hat{x}_k),
          \label{eq:rec}
        \end{equation}
        where $Dis(\cdot)$ refers to a distance function in sample space. Reminding of the optimization objective in Eq. \ref{eq:dmad}, we implement the two constraints using $L_{com}$ and $L_{df}$ respectively. Finally, training is performed by minimizing the overall loss:
        \begin{equation}
          L_{all} = L_{rec} + \gamma _1 L_{com} + \gamma _2 L_{df}.
        \end{equation}

        \textbf{Inference phase.}
        In the inference phase, we use $\mathbf{O}$ and the reconstruction loss to calculate the error maps of input sample $x$:
        \begin{equation}
            A_{rec} = Dis(x, \hat{x}_K),
        \end{equation}
        \begin{equation}
          A_{df} = \sum _k {\Vert O_k \Vert} _2.
          \label{eq:adf}
        \end{equation}
        Image-level anomaly score is computed based on local maxima:
        \begin{equation}
          {Score}_{I} = max(A_{rec} \otimes k^\star) + \alpha max(A_{df} \otimes k^\star).
          \label{eq:det}
        \end{equation}
        where $\otimes$ is convolution operator and $k^\star$ is convolution kernel for anomaly maps. $\alpha$ is a tradeoff parameter. That is, the reconstruction loss and deformation jointly determine the anomaly score, which is much more effective than traditional reconstruction-based methods, as illustrated in \cref{fig:cond}.
            
    \subsection{Variant of PDM}
        \textbf{Modified framework with Pre-PDM.} 
        The DMAD framework together with ICM and PDM modules proposed above are suitable in many anomaly detection scenarios including video surveillance. However, for industrial defect detection, texture reconstruction may be harmful (such as spots on ``Pill''), and we should reconstruct high-level semantic features instead. Since PDM does not work in high-dimensional feature space and interferes with the training process, we propose  variant of PDM, \textit{Pre-PDM (\textbf{PPDM})}, as a solution. PPDM works in sample space, and is applied to the input sample rather than the reconstructed one. \cref{eq:dmad} is modified naturally as follows:
        \begin{equation}
          \begin{split}
          L = & {\Vert x \circ \psi^{\prime}(x) - g(\phi(f(x \circ \psi^{\prime}(x)), z)) \Vert} _2 \\
          & + \gamma _1 R_1(\phi) + \gamma _2 R_2(\psi ^{\prime}).
          \end{split}
          \label{eq:dmad+}
        \end{equation}
        
        Since we do not reconstruct the original samples, reconstruction loss cannot constrain PPDM to maintain information diversity. In order to prevent $x \circ \psi^{\prime}(x)$ in \cref{eq:dmad+} from shortcut learning, we propose to add the inversion of forward deformation, backward deformation $ \mathbf{O}^T$, based on cycle-consistency principle to maintain the diversity of appearance information:
        \begin{equation}
          \psi^{\prime}(x) = Up(h^{\prime}(PE(x)))=\left\{ \mathbf{O},  \mathbf{O}^T\right\}.
        \end{equation}
        
        \textbf{Training phase.}
        The additional cycle-consistency losses $L_{cyc}$ and constraint for the forward-backward deformation $L_{df}^{+}$ are:
        \begin{equation}
          L_{cyc} = {\Vert x - x \circ O_1 \cdots \circ O_K \circ O_K^T \cdots \circ O_1^T\Vert} _2,
        \end{equation}
        \begin{equation}
          L_{df}^{+} = \sum_{k} {\Vert \nabla O_k\Vert} _1 + {\Vert \nabla O^T_k\Vert} _1 + {\Vert O_k\Vert} _2 + {\Vert O^T_k\Vert} _2.
        \end{equation}
        So we train our anomaly detection model with PPDM by minimizing the following loss:
        \begin{equation}
          L_{all}^+ = L_{rec} + \gamma _1 L_{com} + \gamma _2 L_{df}^+ + \gamma _3 L_{cyc}.
        \label{gamma3}
        \end{equation}
        The modified DMAD framework (\ie with PPDM and Reverse Distribution \cite{rd}) is shown in in \cref{fig:3}b. 
        
        \begin{figure}
          \centering
          \includegraphics[width=0.75\linewidth]{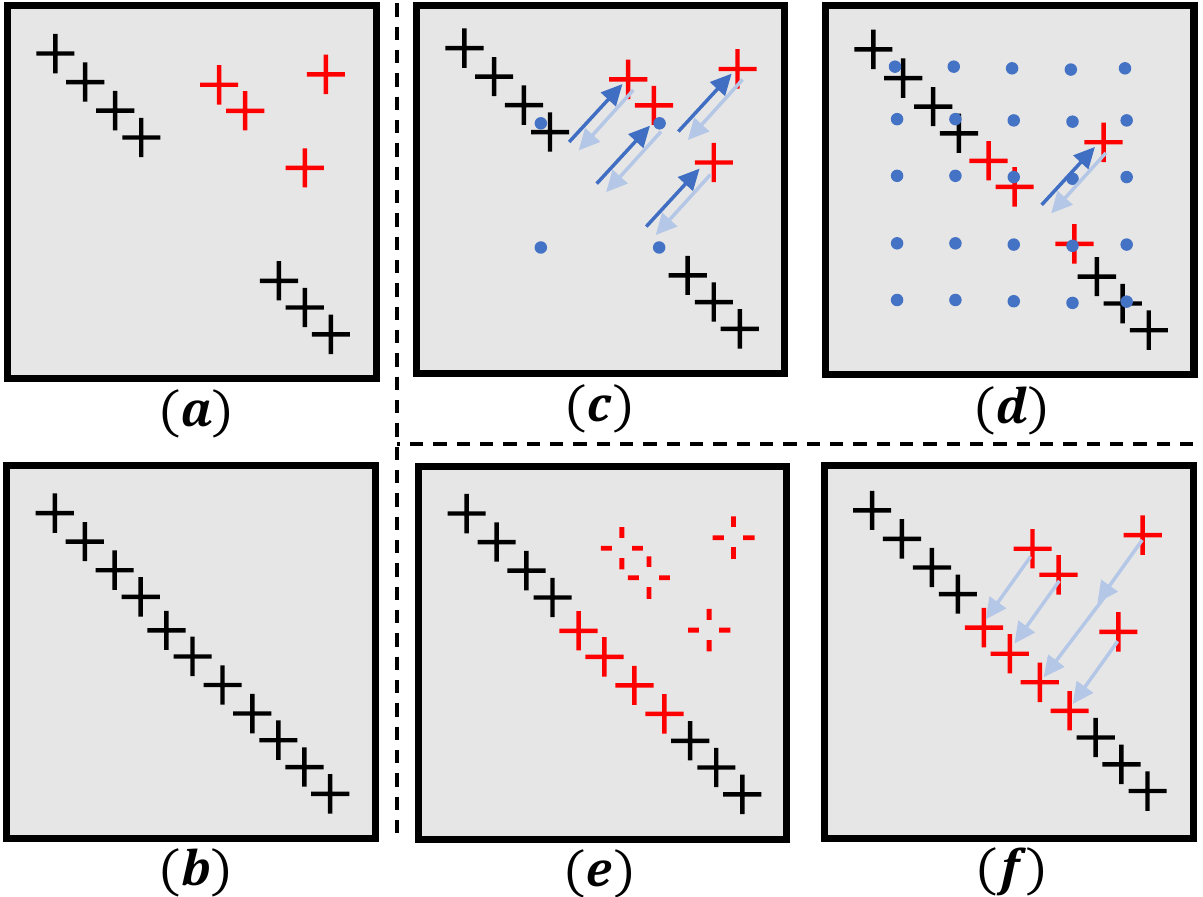}
          \vspace{-0.1em}
          \caption{The process of forward and backward deformation: \textbf{a)} The pattern of original image. Red indicates high error regions and black indicates low error regions; \textbf{b)} The normal pattern; \textbf{c)} The coarse estimation $O_1$ (dark blue) and $O_1^T$ (light blue); \textbf{d)} The fine estimation $O_2$ (dark blue) and $O_2^T$ (light blue); \textbf{e)} Using $\hat{x}$ simply causes high anomaly score at the wrong location (solid cross). Ground-truth is shown as hollow cross; \textbf{f)} $\hat{x} \circ O_2^T \circ O_1^T$ adjusts the high reconstruction error to original position.}
          \label{fig:5}
        \end{figure}
        
        \textbf{Inference phase.}
        The result of PPDM is not aligned with the position of original input, which may reduce the performance of anomaly localization. In order to obtain the anomaly map of real location, we use backward deformation to perform inverse sampling (\cref{fig:5}):
        \begin{equation}
          A_{rec}^{+} = Dis(x, \hat{x}_K) \circ O_K^T \cdots \circ O_1^T,
        \end{equation}
        \begin{equation}
          \begin{split}
          A_{df}^{+} = & \sum _k {\Vert O_k \circ O_{k+1}\cdots \circ O_K\Vert} _2 \\
          + & \sum _k {\Vert O_k^T \circ O_{k-1}^T \cdots \circ O_1^T\Vert} _2. \\
          \end{split}
        \end{equation}
        The image-level anomaly score ${Score}_I^{+}$ and pixel-level anomaly score ${Score}_P^{+}$ are calculated respectively as:
        \begin{equation}
          {Score}_{I}^+ = max(A_{rec}^+ \otimes k^\star) + \alpha max(A_{df}^+ \otimes k^\star),
          \label{eq:det+}
        \end{equation}           
        \begin{equation}
          {Score}_P^{+}  = A_{rec}^{+} + \alpha A_{df}^{+}.
          \label{eq:loc+}
        \end{equation}
            
\section{Experiments and Analysis}
    Firstly, we perform toy experiment on MNIST to illustrate  our approach. Then the quantitative and qualitative results of two versions our DMAD framework are reported in video anomaly detection and industrial surface defect detection respectively. Finally, we conduct ablation experiments and analyze the results.
    \subsection{Datasets}
        \textbf{Surveillance Videos.} Ped2 \cite{ped2}, Avenue \cite{avenue} and ShanghaiTech \cite{timesparse} are fixed-view videos. The anomalies include driving, cycling, running, throwing stuff, etc. Mutual occlusion, anomaly-like behavior, contaminated data and different scenes occur frequently in these datasets.

        \textbf{Industrial Images.} MVTec \cite{mvtec} contains 15 types of industrial images, which are divided into 5 types of textures and 10 types of objects. The defects include crack, scratch, etc. The normal workpiece has different positions, angles and textures. It is used for detection and localization task.
    
    \subsection{Toy experiment}
\begin{figure}[t]
          \centering
          \includegraphics[width=0.85\linewidth]{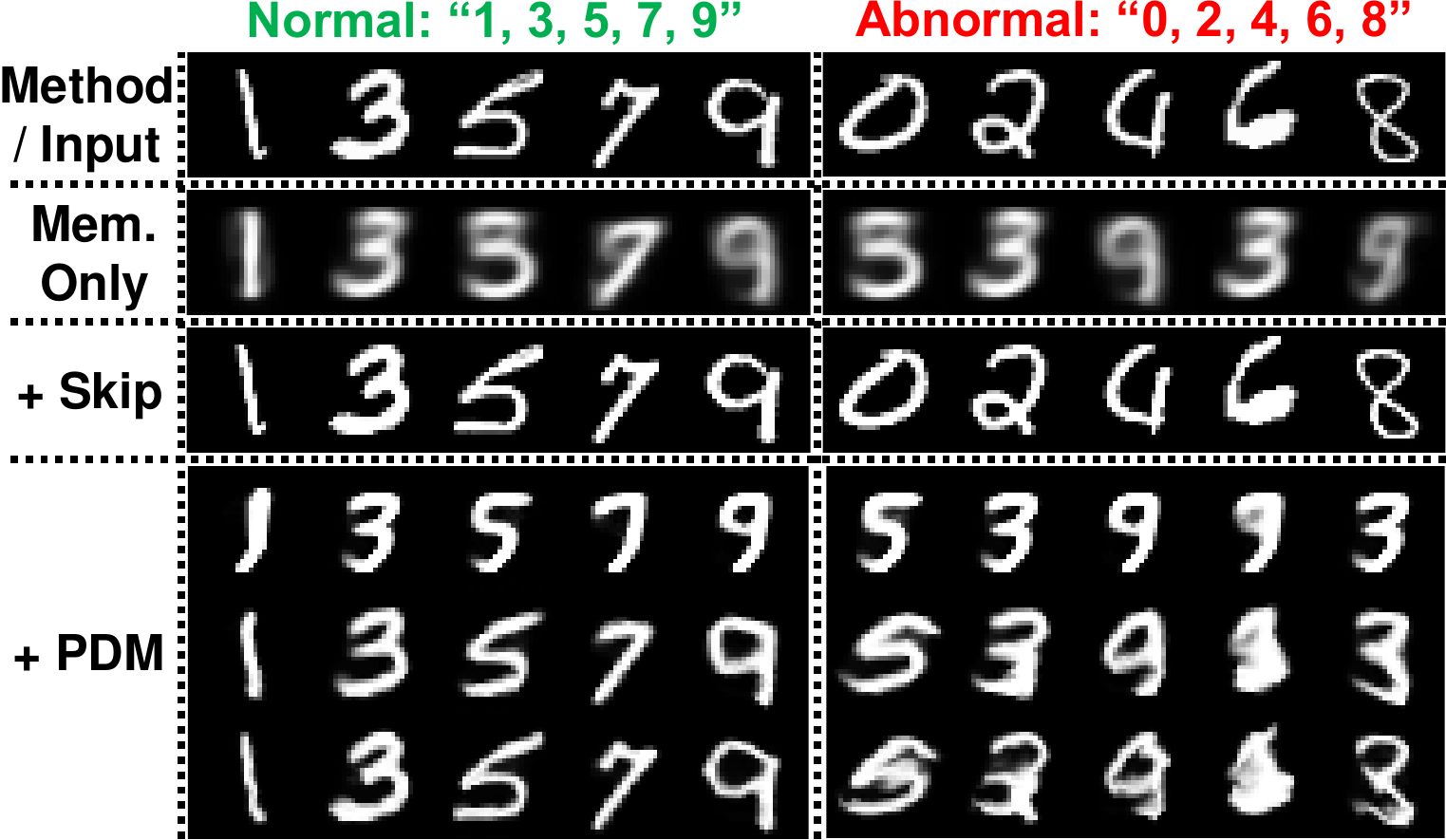}
          \caption{Visualization of toy experiment. The last three rows are: reference reconstructed from memory, reference after coarse deformation, and final output after applying fine deformation.}
          \vspace{-0.1em}
          \label{fig:6}
        \end{figure}

        As shown in \cref{fig:6}, we perform a toy experiment on MNIST dataset \cite{mnist} with setting analogous to out-of-distribution (OOD) detection (\ie training on ``1, 3, 5, 7, 9'' and test on all classes). Our model searches separate memory items for each digital category to reconstruct it into class-specific reference and uses the deformation fields from PDM to adjust it hierarchically. When tested with seen and unseen classes, the model adjusts reconstructed references to normal inputs, but fails on the abnormal ones. 

        In contrast, memory network without diversity-aware module cannot guarantee the intra-class compactness and the reconstruction diversity, which misdirects the model to obtains dataset-optimal ``average memory'', leading to fuzzy reconstruction and lower discrimination ability. The model with full-channel skip-connection suffers from shortcut learning and reconstruct anomalies successfully which weakens the ability to identify anomalies.

    \subsection{Implementation details}
        The input images are resized into $256 \times 256$ and normalized to the value in range of $[-1, 1]$. According to frame-pred strategy \cite{framepred}, the history length is set to 4 in video anomaly detection and 0 for images. After feature extraction with a backbone, PDM and PPDM obtain different heads by stacking stride-2 convolution layers. Then we use $Tanh$ as the activation function of output layer and clipping function to ensure the value of deformation is between $[-1, 1]$. Unless otherwise noted, the architecture of autoencoder complies with the settings of MNAD \cite{mnad} and RD \cite{rd} for PDM and PPDM respectively. The function $Dis(\cdot)$ in \cref{eq:rec} is a combination of $MSE$ loss and $Grad$ loss for reconstruction in sample space, and $COS$ loss for deep features. We set $(\gamma_1, \gamma_2)=(1, 0.25)$ for PDM, $(1, 1)$ for PPDM ($\gamma_3$ is discussed in \cref{ablation}), and $\beta=0.25$ as in \cite{vqvae}. During post-processing, we use average kernel for surveillance videos and Gaussian kernel with $\sigma=4$ for industrial images as in \cite{rd}. Frame-difference method is applied to remove static anomalies for Avenue, because our method detect all anomalies which may be labeled as normal ones in \cite{avenue}. And we use the mask calculated in \cref{eq:mask} to normalize anomaly map in \cite{timesparse} due to the scale changes. Besides, we set $\alpha=0.2$ for three video tasks and an additional option $\alpha=0.05$ for defect detection, depending on whether the defect includes geometrical changes. 
        The model is optimized by AdamW \cite{adamw} and the learning rate is 2e-4 and 5e-3 as in \cite{mnad,rd} respectively, being decayed by CosineAnnealingLR\cite{coslr} strategy. We adopt 60, 60, 60, 10, 400 epochs for MNIST, Ped2, Avenue, Shanghai and MVTec respectively with the batch size of 8. 
    
    \subsection{Main results}
        \textbf{Surveillance Videos.}
        We compare our method with SOTA works on video anomaly detection in \cref{tab:1}. Our method outperforms comparative approaches even though we neither use external estimators nor remove abnormal frames in training data. In additional, if we detect global offset for camera jitter in \cite{avenue}, there will be extra $0.1\%$ gain.
        
\begin{figure}[t]
      \centering
      \includegraphics[width=0.56\linewidth]{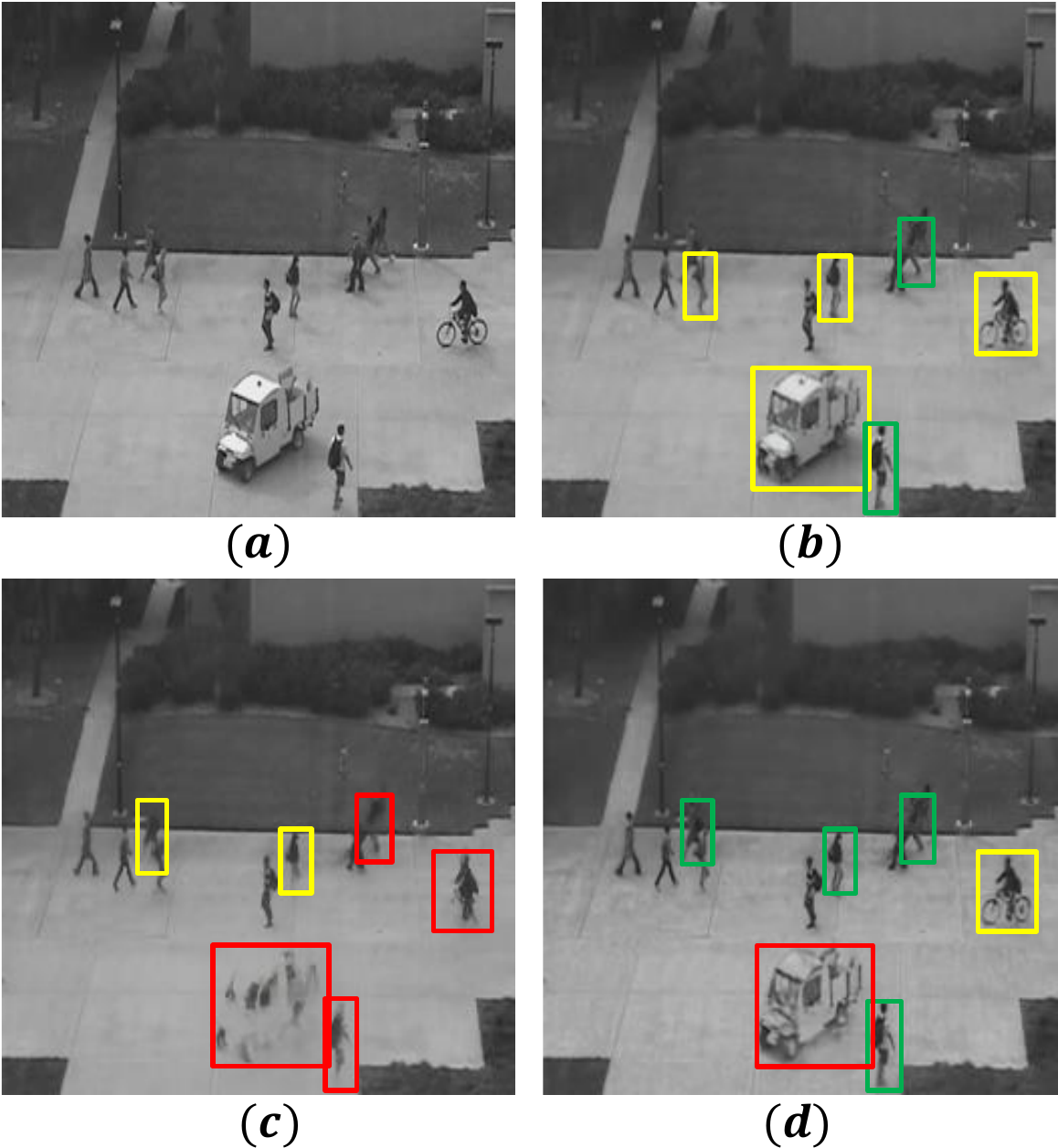}
      \caption{Qualitative results on Ped2. \textbf{a)} Original image; \textbf{b)} Memory network with skip-connection \cite{mnad}; \textbf{c)} PDM with VQ-Layer \cite{vqvae}; \textbf{d)} PDM with ICM. Green box indicates negative, red box indicates positive, and yellow box indicates weak positive.}
      \label{fig:7}
      \vspace{-0.1em}
    \end{figure}

        The qualitative results are shown in \cref{fig:7}. We find that: anomalies are over-reconstructed in (b); normals are not reconstructed well in (c); (d) greatly improves normal reconstruction with slightly less inhibition ability for anomalies. 

        \begin{table}
        \renewcommand\arraystretch{0.8}
          \centering
          \caption{
            Video anomaly detection results on Ped2\cite{ped2}, Avenue\cite{avenue} and Shanghai\cite{timesparse}. We calculate $AUC(\%)$ with all frames together. Numbers in bold indicate the best performance and the underlined ones are the second best. $^+$ indicates that we reproduce the result due to higher performance or absence of implementation. 
         }
      \begin{tabular}{@{}c|ccc@{}}
        \toprule
        \makebox[0.12\textwidth][c]{Methods} & 
        \makebox[0.075\textwidth][c]{Ped2\cite{ped2}} & 
        \makebox[0.08\textwidth][c]{Avenue\cite{avenue}} & 
        \makebox[0.115\textwidth][c]{Shanghai\cite{timesparse}} \\
        \midrule
        Conv2D\cite{convae2d}            & 90.0      & 70.2  &  - \\
        Conv3D\cite{convae3d}            & 91.2      & 77.1  &  - \\
        ConvLSTM\cite{convaelstm}        & 88.1      & 77.0  &  - \\
        FramePred\cite{framepred}          & 95.4      & 84.9  &  72.8 \\
        \midrule
        ConvVQ$^+$        & 90.2      & 84.3   &  -  \\
        MemAE\cite{memae}                & 94.1      & 83.3  & 71.2  \\
        MPN\cite{mpn}                  & 96.9      & 89.5  &  73.8  \\
        MNAD$^+$\cite{mnad}             & 97.8      & 88.5  &  70.5  \\
        HF$^2$VAD\cite{hf2vad}           & \uline{99.3}      & \uline{91.1}   & \uline{76.2}  \\
        \midrule
        Ours  & \textbf{99.7} & \textbf{92.8}  & \textbf{78.8}  \\
        \bottomrule
      \end{tabular}
      \label{tab:1}
    \end{table}
        
    \begin{table*}
      \centering
      \renewcommand\arraystretch{0.7}
      \caption{Image-level $AUC(\%)$ of anomaly detection on MVTec\cite{mvtec}.
      }
      \setlength{\tabcolsep}{1mm}{
      \begin{tabular*}{\linewidth}{@{}c|cccccccc|c@{}}
        \toprule
        \makebox[0.12\textwidth][c]{Class$\backslash$Methods} & \makebox[0.06\textwidth][c]{GN\cite{gn}} & \makebox[0.095\textwidth][c]{PSVDD\cite{psvdd}} & \makebox[0.09\textwidth][c]{DAAD\cite{daad}} & \makebox[0.1\textwidth][c]{CutPaste\cite{cutpaste}} & \makebox[0.08\textwidth][c]{PaDiM\cite{padim}} & \makebox[0.11\textwidth][c]{PatchCore\cite{patchcore}} & \makebox[0.10\textwidth][c]{DRAEM\cite{draem}} & \makebox[0.07\textwidth][c]{RD\cite{rd}} & \makebox[0.07\textwidth][c]{Ours} \\
        \midrule
        Avg$_{Textures}$ & 77.5 & 94.5 & 91.0 & 97.5 & 98.8 & 99.0 & 99.1 & \uline{99.5} & \textbf{99.9}\\
        \midrule
        Avg$_{Objects}$ & 75.5 & 90.8 & 88.8 & 95.5 & 93.8 & \uline{99.2} & 97.4 & 98.0 & \textbf{99.3}\\
        \midrule
        Avg$_{All}$ & 76.2 & 92.1 & 89.5 & 96.1 & 95.5 & \uline{99.1} & 98.0 & 98.5 & \textbf{99.5}\\
        \bottomrule
      \end{tabular*}}
      \label{tab:3}
    \end{table*}
        
    \begin{table*}
        \renewcommand\arraystretch{0.7}
          \centering
          \caption{Pixel-level $AUC(\%)$ of anomaly localization on MVTec\cite{mvtec}.
          }
          \setlength{\tabcolsep}{1mm}{
          \begin{tabular*}{\linewidth}{@{}c|ccccccc|c@{}}
            \toprule
            \makebox[0.13\textwidth][c]{Class$\backslash$Methods} & \makebox[0.11\textwidth][c]{PSVDD\cite{psvdd}} & \makebox[0.1\textwidth][c]{CutPaste\cite{cutpaste}} & \makebox[0.1\textwidth][c]{PaDiM\cite{padim}} & \makebox[0.11\textwidth][c]{PatchCore\cite{patchcore}} & \makebox[0.11\textwidth][c]{DRAEM\cite{draem}} &  \makebox[0.09\textwidth][c]{TMAE\cite{tmae}} &\makebox[0.08\textwidth][c]{RD\cite{rd}} & \makebox[0.08\textwidth][c]{Ours} \\
            \midrule
            Avg$_{Textures}$ & 93.7 & 96.3 & 96.9 & 97.6 & \textbf{97.9} & 93.8 & 97.7 & \uline{97.8}\\
            \midrule
            Avg$_{Objects}$ & 96.7 & 95.8 & 97.8 & \textbf{98.4} & 97.0 & 94.0 & 97.9  & \uline{98.3}\\
            \midrule
            Avg$_{All}$ & 95.7 & 96.0 & 97.5 & \uline{98.1} & 97.3 & 93.9 & 97.8 & \textbf{98.2}\\
            \bottomrule
          \end{tabular*}}
          \vspace{-0.5em}
          \label{tab:4}
        \end{table*}
        
        \textbf{Industrial Images.}
        Anomaly detection results on MVTec are shown in \cref{tab:3} and localization results are shown in \cref{tab:4}. With the deformation from PPDM, texture anomalies are detected with high performance and our method outperforms SOTA methods in both detection and localization tasks without memorizing an enormous number of embedding from training data.
        
        \subsection{Ablation study}
        \label{ablation}
        As shown in \cref{tab:2}, single-output memory module without PDM suppresses diverse normal patterns seriously, 
        while separate PDM without memory provides comparable performance gains as the previous SOTA works because the module ``\textit{Comp.}" serves as an information compression module instead. The number of multi-scale deformation fields also have a modest effect on performance. We suggest that ``\textit{$K$}" should at least make control grid scale cover the size of base elements (\eg pedestrian limbs). Besides, the foreground-background selection module further improves the compactness of memory embedding.
        Moreover, if any constraint for PDM is missing, the abnormal information will be transmitted and cause shortcut learning. Especially, the cycle-consistency constraint $L_{cyc}$ is also a necessary part for PPDM to avoid degenerated solution ($-1.7\%$), because the feature reconstruction error can be minimized by eliminating all necessary information.
        
        \begin{table}
        \renewcommand\arraystretch{0.8}
         \centering
          \caption{Ablation study of proposed module and loss on Ped2\cite{ped2}.
          From left to right: without PDM; with only one head of PDM ($K$=1); without memory module; without background template; removing strength constraint; removing smoothness constraint.
          }
          \begin{tabular*}{\linewidth}{@{}c|cccc|cc@{}}
            \toprule
            \textit{w / o}&\makebox[0.04\textwidth][c]{PDM} & \makebox[0.04\textwidth][c]{$K>1$} & \makebox[0.04\textwidth][c]{Mem} & \makebox[0.04\textwidth][c]{$x_{bg}$} & \makebox[0.04\textwidth][c]{${\Vert O \Vert} _2$} & \makebox[0.04\textwidth][c]{${\Vert \nabla O \Vert} _2$}\\
            \midrule
            \makebox[0.065\textwidth][c]{$AUC\%$} & -9.4 & -0.4 & -1.8 & -1.6 & -1.4 & -1.4\\
            \bottomrule
          \end{tabular*}
          \label{tab:2}
        \end{table}
        
        \begin{table}
    \renewcommand\arraystretch{0.8}
      \centering
      \caption{
        Ablation study of deformation constraint on MVTec\cite{mvtec}.
     }
      \begin{tabular}{c|cccccc}
        \toprule
        \makebox[0.07\textwidth][c]{Task $\backslash$ $\gamma_3$} & \makebox[0.04\textwidth][c]{0} &\makebox[0.04\textwidth][c]{0.1} & \makebox[0.04\textwidth][c]{0.25} & \makebox[0.04\textwidth][c]{0.5} & \makebox[0.04\textwidth][c]{1} & \makebox[0.04\textwidth][c]{2} \\
        \midrule
        Det.  & 97.9 & 99.4  & 99.4 &   \uline{99.5}   & \textbf{99.6} &  99.4\\
        \midrule
        Loc.  &  96.6 & \uline{98.0} & \textbf{98.2} &  \uline{98.0}  & 97.7  & 97.6 \\
        \bottomrule
      \end{tabular}
      \label{tab:5}
    \end{table}
        
        As shown in \cref{tab:5}, the proposed method is robust to hyperparameter $\gamma_3$ in \cref{gamma3}. 
        On condition that degenerated solution will not be formed, weakening the constraint makes model restore images from reference with less detail more easily and perceive the position of anomalies more accurately by transforming anomalies to normal patterns which benefits localization task. On the contrary, strengthening the constraint alleviates shortcut learning and improves image-level result by maintaining more abnormal details.
        
    \subsection{Discussion}
        \textbf{Contaminated data assumption.}
        Assuming that training data only contains completely normal data is unrealistic, because the workload of natural data cleansing is considerable, even has same cost as data annotations. We mix training data with abnormal events in Ped2 to simulate contaminated data and find performance drop of MNAD\cite{mnad} is $-3.7\%$ while ours is $-1.8\%$. The reason why DMAD is less affected may be that PDM can encode and transmit residual abnormal representation of mixed anomalies and anomaly-like normal samples, thus strictly maintaining representation compactness of the main encoder and low generalization ability on abnormal samples.
        
\begin{figure}[t]
          \centering
          \includegraphics[width=0.85\linewidth]{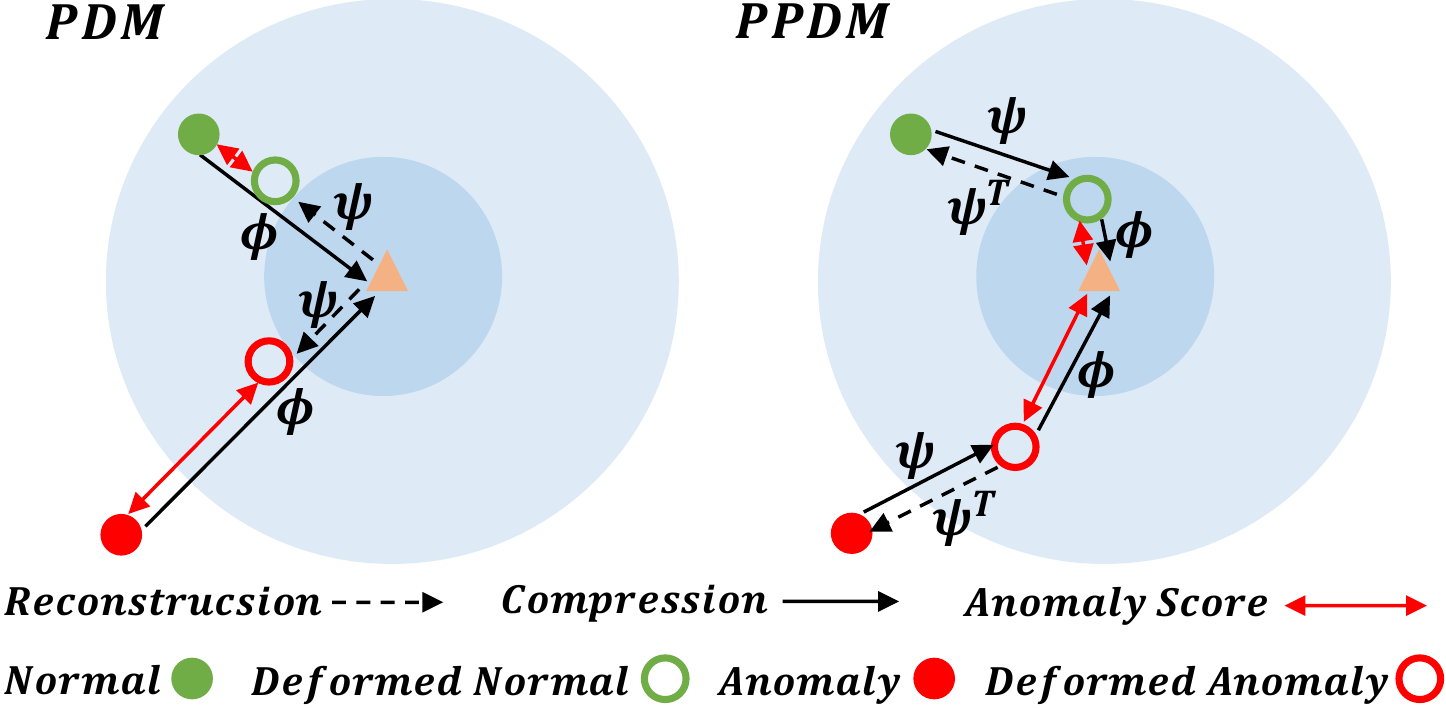}
          \caption{Illustration of two DMAD implementations with PDM and PPDM respectively.
          }
          \label{fig:dis}
          \vspace{-0.1em}
        \end{figure}

        \textbf{Two DMAD implementations.}
        As introduced earlier, our DMAD framework is implemented by two versions using PDM and PPDM respectively, corresponding to different detection targets.
        As shown in \cref{fig:dis}, PDM learns measurable quantization error (caused by ICM) from memory embedding to diverse patterns which enhances the potential of ICM to keep intra-class compactness without generating unmeasureable reconstruction loss. Unlike PDM, PPDM is partially in charge of the information compression, \ie PPDM removes the diverse appearance by using the reverse deformation process from inputs to reference.
        
\section{Conclusion}
    In this paper, we present a reconstruction-based diversity-measurable anomaly detection framework, which simultaneously enhances anomaly discrimination and reconstruction diversity. Pyramid deformation module is proposed to be used together with information compression module for this purpose. PDM models multi-scale transformation fields from reference to original input explicitly without relying on external estimators. Therefore, diverse normal patterns can be reconstructed and anomaly severity can be measured accurately. Empirical studies on both videos and images benchmarks show the effectiveness and applicability of our work. In future research, we will further explore diversity-aware models for anomaly detection.
    
    \paragraph{Limitations.}
    Our method focuses on anomaly with measurable geometrical diversity, the most common type in anomaly detection. 
    However, as for anomaly with other kind of diversities, \eg colors, the proposed diversity measure may not be positively correlated to anomaly severity.

{\small
\bibliographystyle{ieee_fullname}
\bibliography{egbib}
}

\end{document}